\documentclass[11pt]{article}
\PassOptionsToPackage{table}{xcolor}
\usepackage{amsmath}
\usepackage{amssymb}
\usepackage{algorithm}
\usepackage{algpseudocode}
\algrenewcommand\algorithmicrequire{\textbf{Require:}}
\algrenewcommand\algorithmicensure{\textbf{Ensure:}}
\usepackage[final]{acl}
\usepackage{booktabs}
\usepackage{multirow}

\usepackage{times}
\usepackage{latexsym}

\algrenewcommand\algorithmicindent{1.1em}

\algrenewcommand\algorithmiccomment[1]{\hfill{\footnotesize$\triangleright$~#1}}

\usepackage[T1]{fontenc}

\usepackage[utf8]{inputenc}

\usepackage{microtype}

\usepackage{inconsolata}

\usepackage{graphicx}
\usepackage{placeins}

\usepackage[most]{tcolorbox}
\newtcolorbox{casebox}[1]{
  enhanced,
  colback=green!3!white,
  colframe=green!50!black,
  fonttitle=\bfseries,
  title=#1,
  boxrule=0.8pt,
  arc=2pt,
  left=6pt, right=6pt, top=4pt, bottom=4pt,
  before skip=8pt, after skip=8pt
}

\title{Reasoning Fails Where Step Flow Breaks}

\author{
  \textbf{Xiaoyu Xu\textsuperscript{1,2}},
  \textbf{Yulan Pan\textsuperscript{2}},
  \textbf{Xiaosong Yuan\textsuperscript{3}}$^\ddagger$,\\
  \textbf{Zhihong Shen\textsuperscript{2}},
  \textbf{Minghao Su\textsuperscript{2}},
  \textbf{Yuanhao Su\textsuperscript{2}},
  \textbf{Xiaofeng Zhang\textsuperscript{1}$^\dagger$}
\\
\\
  \textsuperscript{1}Shanghai Jiao Tong University,
  \textsuperscript{2}Fuzhou University,
  \textsuperscript{3}Jilin University
\\
     {\tt\small framebreak@sjtu.edu.cn}\quad
}

\begin{document}
\maketitle

\begin{abstract}
Large reasoning models (LRMs) that generate long chains of thought now perform well on multi-step math, science, and coding tasks. However, their behavior is still unstable and hard to interpret, and existing analysis tools struggle with such long, structured reasoning traces. We introduce \emph{Step-Saliency}, which pools attention--gradient scores into step-to-step maps along the question--thinking--summary trajectory. Across several models, Step-Saliency reveals two recurring information-flow failures: \textbf{Shallow Lock-in}, where shallow layers over-focus on the current step and barely use earlier context, and \textbf{Deep Decay}, where deep layers gradually lose saliency on the thinking segment and the summary increasingly attends to itself and the last few steps. Motivated by these patterns, we propose \emph{StepFlow}, a saliency-inspired test-time intervention that adjusts shallow saliency patterns measured by Step-Saliency via Odds-Equal Bridge and adds a small step-level residual in deep layers via Step Momentum Injection. StepFlow improves accuracy on math, science, and coding tasks across multiple LRMs without retraining, indicating that repairing information flow can recover part of their missing reasoning performance. Code is available at \url{https://github.com/XiaoyuXu-Vincent/step-saliency}.

\begingroup
\renewcommand\thefootnote{}
\footnotetext{${}^\dagger$ Corresponding author,${}^\ddagger$ Project leader}
\endgroup

\end{abstract}

\section{Introduction}

Large reasoning models (LRMs) generate an internal thinking segment before producing a final summary, forming a question--thinking--summary trajectory. These models achieve strong performance on multi-step mathematics, code generation, and scientific question answering~\cite{ahn2024large,plaat2025multi,xu2025towards}. Despite these gains, LRMs still exhibit persistent failure modes, including unfaithful chains of thought~\cite{paul2024making,lightman2023let,barez2025chain}, overconfident hallucinations~\cite{fan2025improving,cao2025towards}, and brittleness on compositional tasks~\cite{song2025survey,li2025system,boye2025large,simhi2025trust,chen2025reasoning}. When an LRM makes a mistake, we lack a clear way to attribute the final error to the model's internal reasoning trace.
This motivates a diagnostic that tracks step-to-step influence across depth and links shifts in that influence to incorrect final answers~\cite{luo2024understanding,cambria2024xai,yangdemystifying}.

\begin{figure*}[t]
  \centering
  \includegraphics[width=\linewidth]{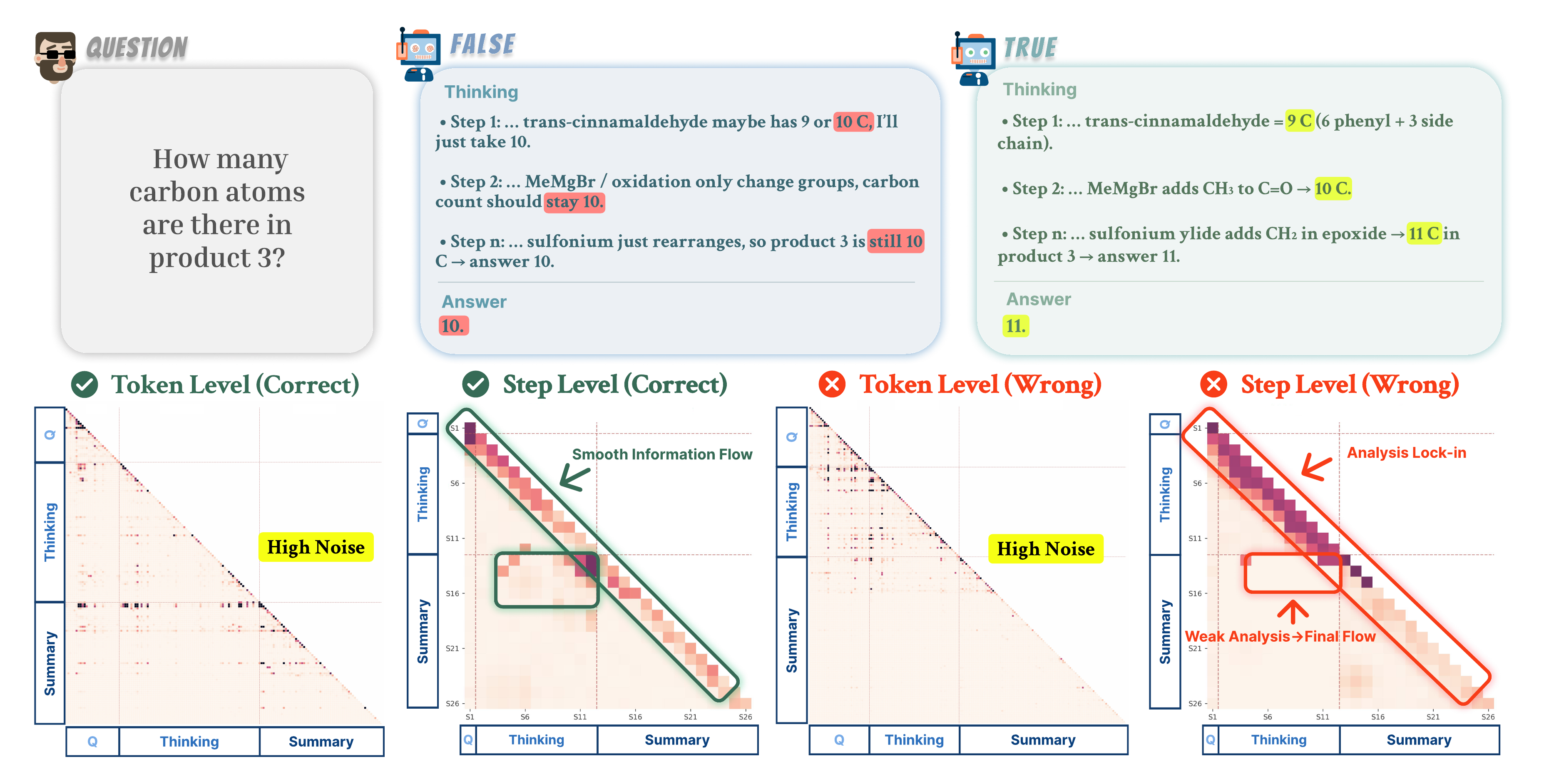}
  \caption{
  \textbf{From token- to step-level saliency.}
  Token-level saliency maps are dense and noisy; Step-Saliency pools them into question/thinking/summary blocks.
  Correct traces show smooth question$\rightarrow$thinking$\rightarrow$summary flow, while errors exhibit shallow lock-in and weak thinking$\rightarrow$summary links.
  }
  \label{fig:step-saliency-case}
\end{figure*}

Prior work offers several token-level diagnostics for inspecting LRMs during generation.
Attention-based analyses visualize which tokens receive attention when producing later tokens~\cite{vig2019analyzing,yeh2023attentionviz}, but attention weights are not always faithful indicators of what actually drives the prediction~\cite{jain2019attention}.
Saliency-based methods instead assign gradient-weighted importance to earlier tokens~\cite{ancona2017towards,wu2023analyzing}, yet along long reasoning traces the scores can be noisy and difficult to aggregate across positions.
The main difficulty is not the lack of signals but the lack of a readable unit aligned with reasoning steps: token-level maps are dense, local, and do not naturally summarize step-to-step dependence.
As Figure~\ref{fig:step-saliency-case} shows, even when correct and error traces differ, the signal is dispersed across many tokens, making it difficult to summarize which steps remain influential later and how this changes with depth.

To address this limitation, we introduce \textbf{Step-Saliency}, a step-level diagnostic for long-form reasoning.
We compute token saliency using attention--gradient influence and then pool the scores within each step.
This yields a compact step$\to$step map that is easy to compare across layers.
Applying Step-Saliency to several LRMs, we observe two common patterns in incorrect outputs.
In shallow layers, influence concentrates on the current thinking step and its immediate neighbors, while earlier context is suppressed; we call this \textbf{Shallow Lock-in}.
As depth increases, saliency on the thinking segment weakens for both correct and incorrect outputs, but it fades much faster on incorrect ones, and the summary becomes more self-focused; we call this \textbf{Deep Decay} (\S\ref{sec:maps-observe}).

Building on these observations, we propose \textbf{StepFlow}, a lightweight test-time intervention that modifies a model's forward pass without retraining or backpropagation.
StepFlow has two components that target the two failure patterns.
\textbf{Odds-Equal Bridge (OEB)} is applied to the first few layers: when the model is generating a thinking step, it prevents step-level mass from collapsing onto the just-written tokens by keeping a non-trivial share on the question and earlier steps.
\textbf{Step Momentum Injection (SMI)} is applied to the last few layers: at step boundaries, it carries a small residual summary of the previous step into the next step, so earlier reasoning remains available when the summary is produced.
Together, OEB and SMI promote a steadier question--thinking--summary linkage during generation.

We evaluate StepFlow on DeepSeek-R1-Distill (7B/14B/32B), GPT-OSS-20B, and QwQ-32B-Preview across six benchmarks: AIME24, AIME25, AMC23, MATH-500, GPQA-Diamond, and LiveCodeBench. StepFlow consistently improves accuracy across all backbones, with the largest gains on competition-style problems that require long reasoning chains. Our main contributions are as follows:

\begin{itemize}
\item We introduce \textbf{Step-Saliency}, a diagnostic that aggregates 
  token-level saliency into step$\to$step maps, making long reasoning 
  traces interpretable at the step level.
  
\item We identify two depth-wise information-flow failure patterns 
  in large reasoning models, which we term \textbf{Shallow Lock-in} and 
  \textbf{Deep Decay}; they reliably separate incorrect from correct traces 
  in our saliency analysis.
  
\item We propose \textbf{StepFlow}, a test-time intervention that 
  reshapes information flow via Odds-Equal Bridge and Step Momentum 
  Injection, improving accuracy across multiple benchmarks without 
  retraining.
\end{itemize}

\begin{figure*}[t]
  \centering
  \includegraphics[width=\linewidth]{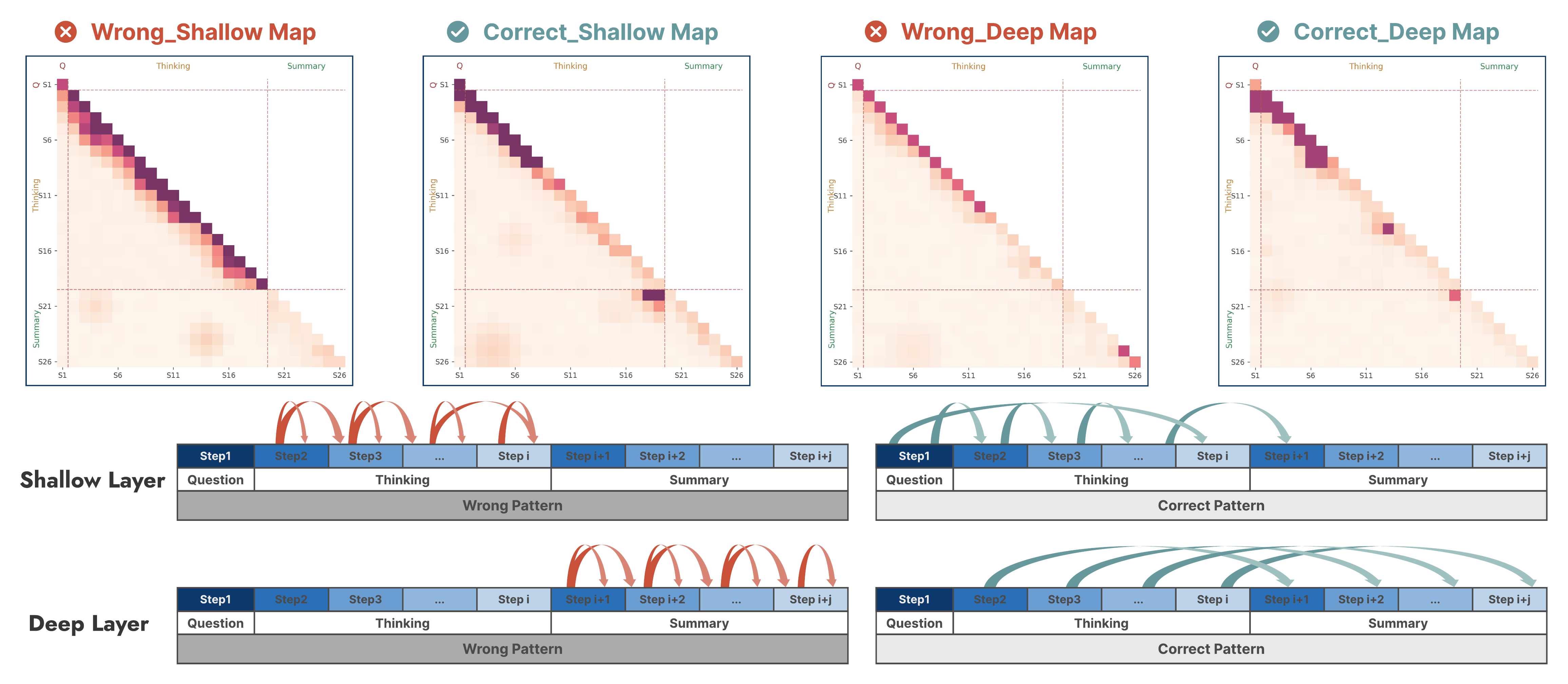}
  \caption{
  \textbf{Step-Saliency patterns for shallow vs.\ deep layers and correct vs.\ error traces.}
  Top: depth-collapsed step$\to$step saliency maps; darker red indicates stronger influence between steps.
  Bottom: schematic diagrams summarizing the observed patterns---red arrows denote narrow, local flow (\emph{Shallow Lock-in} and \emph{Deep Decay} in error traces), while blue arrows denote broad, long-range flow in correct traces.
  }
  \label{fig:saliency-grid}
\end{figure*}

\section{Related Work}

\subsection{Large Reasoning Models}
Chain-of-thought prompting (CoT)~\cite{wei2022chain} substantially improved LLM reasoning and spurred \emph{Large Reasoning Models} (LRMs) with strong multi-step reasoning and structured intermediate traces. Although LRMs demonstrate strong performance on multi-step reasoning benchmarks~\cite{kojima2022large,chen2025towards,xu2025towards,lanham2023measuring}, their reasoning processes still suffer from instability and limited interpretability. Beyond standard CoT prompting, test-time scaling methods such as self-consistency and sampled-path selection improve robustness~\cite{wang2023self,muennighoff2025s1}. Other work builds multi-step frameworks with tools or self-improvement~\cite{li2025start,li2023making,yuan2024instance,dutta2024think}. Our understanding of how information is propagated, transformed, and possibly lost inside LRMs remains incomplete, which makes it difficult to systematically analyze reasoning failures, incorrect inferences, and inconsistent answers~\cite{song2025survey,li2025system,chen2025reasoning}.

\subsection{Layer-wise Reasoning Behavior}
Many studies employing layer-wise analysis have revealed that Transformer-based LLMs exhibit a clear functional division across network depth during reasoning~\cite{geva2021transformer}. In the shallow layers, the model primarily performs local and formal pattern recognition, such as extracting n-gram features~\cite{sajjad2022analyzing,vulic2020probing} and handling structural information~\cite{van2019does}. In the deep layers, the model shifts toward cross-sentence and cross-concept integration, handling semantic alignment~\cite{saglam2025large}, long-range dependency modeling~\cite{wang2020cluster}, and the global organization of the reasoning chain~\cite{dong2023survey}.

\subsection{Information Flow}

To study how information moves inside LLMs, prior work has proposed a range of information-flow analyses~\cite{zhao2024explainability,luo2024understanding,cambria2024xai}.  

\textbf{Attention-based methods} inspect attention weights or head activations~\cite{vig2019analyzing,yeh2023attentionviz} to show which tokens the model focuses on. They are often used to infer context-dependency structures~\cite{vashishth2019attention}. However, attention is not a faithful explanation: different attention maps can lead to the same prediction~\cite{jain2019attention}. Kobayashi et al.~\cite{kobayashi2020attention} therefore separate the contribution of attention weights from that of transformed value vectors, partially decoupling "where the model looks" from "what representation is propagated". Yan et al.~\cite{yan2025don} propose attention-level interventions to better preserve CoT context, complementary to saliency-based analyses.

\textbf{Saliency-based methods} instead assign importance scores to tokens using gradient-based, perturbation-based, or gradient--activation techniques~\cite{ancona2017towards,huber2022benchmarking,ding2021evaluating,boggust2023saliency}. These scores highlight which tokens support answers or intermediate reasoning steps~\cite{wang2020gradient,dong2023survey,wu2023analyzing,lee2025evaluating} and are more tightly coupled to the model's actual computation than raw attention. Yet they are typically local to a single prediction and often noisy, which makes it difficult to obtain stable, sequence-level explanations of how information flows through a long reasoning trace.

\section{Motivation and Information Flow Analysis}

\subsection{Saliency and step-level aggregation}

\paragraph{Saliency (token$\to$token).}
Let $x_{1:T}$ be the generated tokens and $\mathcal{L}_t = -\log p_\theta(x_t \mid x_{<t})$ the token-level loss.
For layer $\ell$ and head $h$, let $A^{(\ell,h)} \in \mathbb{R}^{T \times T}$ denote the causal attention matrix.
For each query position $t$ and key position $k \le t$, we compute the gradient-weighted influence and average over $H$ heads:
\begin{equation}
\label{eq:attn-grad}
I^{(\ell)}_{t\leftarrow k} \;=\; \frac{1}{H}\sum_{h=1}^{H}
\Bigl|\,A^{(\ell,h)}_{t,k} \cdot \frac{\partial \mathcal{L}_t}{\partial A^{(\ell,h)}_{t,k}}\,\Bigr|,
\qquad k \le t.
\end{equation}
We take absolute values to measure influence magnitude regardless of direction; signed scores would complicate step-level aggregation in Eq.~\eqref{eq:step-map} since positive and negative contributions could cancel within a step.\footnote{Signed influence could reveal suppression patterns but would require separate positive/negative pooling, which we leave for future work.}
Stacking over all $t$ gives the full influence matrix $\mathbf{I}^{(\ell)} \in \mathbb{R}^{T \times T}$.
We then apply per-query row normalization to obtain a token$\to$token saliency distribution:
\begin{equation}
\label{eq:row-norm}
\tilde s^{(\ell)}_{t\leftarrow k} \;=\;
\frac{\bigl(\mathbf{I}^{(\ell)}\bigr)_{t\leftarrow k}}
     {\sum_{k'\le t}\bigl(\mathbf{I}^{(\ell)}\bigr)_{t\leftarrow k'}+\varepsilon}\,.
\end{equation}
Compared with raw attention, these scores reflect how much each past token contributes to the loss at position $t$ and remain informative even when attention is diffuse (full computation details in Appendix~\ref{app:saliency-computation}).

\paragraph{From tokens to steps.}
For each generation trace, we segment the sequence into a question segment, a multi-step thinking segment, and a summary segment.
Let $\Gamma_1,\dots,\Gamma_K$ denote the $K$ thinking steps (each $\Gamma_i$ is a contiguous token span) and $\Gamma_{K+1}$ the summary segment; we use $\Gamma_i$ generically to refer to any segment when the context is clear.
For layer $\ell$, we aggregate token-level saliency into step$\to$step blocks:
\begin{equation}
\label{eq:step-map}
M^{(\ell)}_{j\leftarrow i} \;=\;
\frac{1}{|\Gamma_j|\,|\Gamma_i|}
\sum_{t\in \Gamma_j}\sum_{k\in \Gamma_i}
\tilde s^{(\ell)}_{t\leftarrow k},
\qquad i\le j\,.
\end{equation}
We use mean pooling to suppress token-level noise (cf.\ Figure~\ref{fig:step-saliency-case}); the trade-off is some loss of fine-grained signal.
Mass on the block diagonal captures within-segment self-reinforcement; off-diagonal mass captures cross-step and cross-segment flow.
Averaging $M^{(\ell)}$ over layers yields a depth-collapsed step$\to$step map (i.e., averaged over depth), which we refer to as the \emph{Step-Saliency map}.

\subsection{Information flow among steps}
\label{sec:maps-observe}

We compare correct and error traces for identical prompts using layerwise Step-Saliency maps (Figure~\ref{fig:saliency-grid}).
Each map summarizes how strongly a given thinking step or the summary depends on earlier steps across depth.

\paragraph{Shallow layers: Shallow Lock-in.}
In shallow layers, correct traces keep a visible link to the question and spread saliency across several thinking steps.
The model continues to refer back to the problem statement and to earlier parts of its reasoning.
Error traces instead place most saliency inside a narrow band around the current step and its immediate neighbors, while saliency on the question and early thinking steps is strongly suppressed.
The model attends mainly to what it just wrote.
We refer to this local feedback pattern as \textbf{Shallow Lock-in}.

\paragraph{Deep layers: Deep Decay.}
In deeper layers, saliency on the thinking segment decreases for both correct and error traces, but the decay is faster for errors.
For correct traces, summary tokens still allocate noticeable saliency back to several earlier thinking steps, indicating that the answer remains connected to the internal reasoning.
For error traces, deep layers place most saliency on summary tokens themselves and on the last few thinking steps, with very little mass on earlier steps.
We call this faster loss of thinking saliency in deep layers \textbf{Deep Decay}: the summary is produced with only a thin connection to the full reasoning chain.

\subsection{Quantifying inter-step flow}
\label{sec:metrics}

To summarize these map-level patterns, we define two simple layerwise metrics based on the Step-Saliency blocks.

\paragraph{Definitions.}
Let $M^{(\ell)}_{j\leftarrow i}$ be the step$\to$step map at layer $\ell$ from Eq.~\eqref{eq:step-map}.
We summarize within-thinking and within-summary self-reinforcement as
\begin{equation}
I^{(\ell)}_{\mathrm T} = \frac{1}{K}\sum_{i=1}^{K} M^{(\ell)}_{i\leftarrow i}
\quad\text{and}\quad
I^{(\ell)}_{\mathrm S} = M^{(\ell)}_{(K+1)\leftarrow (K+1)},
\label{eq:IT-IS}
\end{equation}
where $I^{(\ell)}_{\mathrm T}$ measures how much each step reuses its own content and $I^{(\ell)}_{\mathrm S}$ measures how much the summary attends to itself across depth.

\begin{figure}[t]
  \centering
  \includegraphics[width=\linewidth]{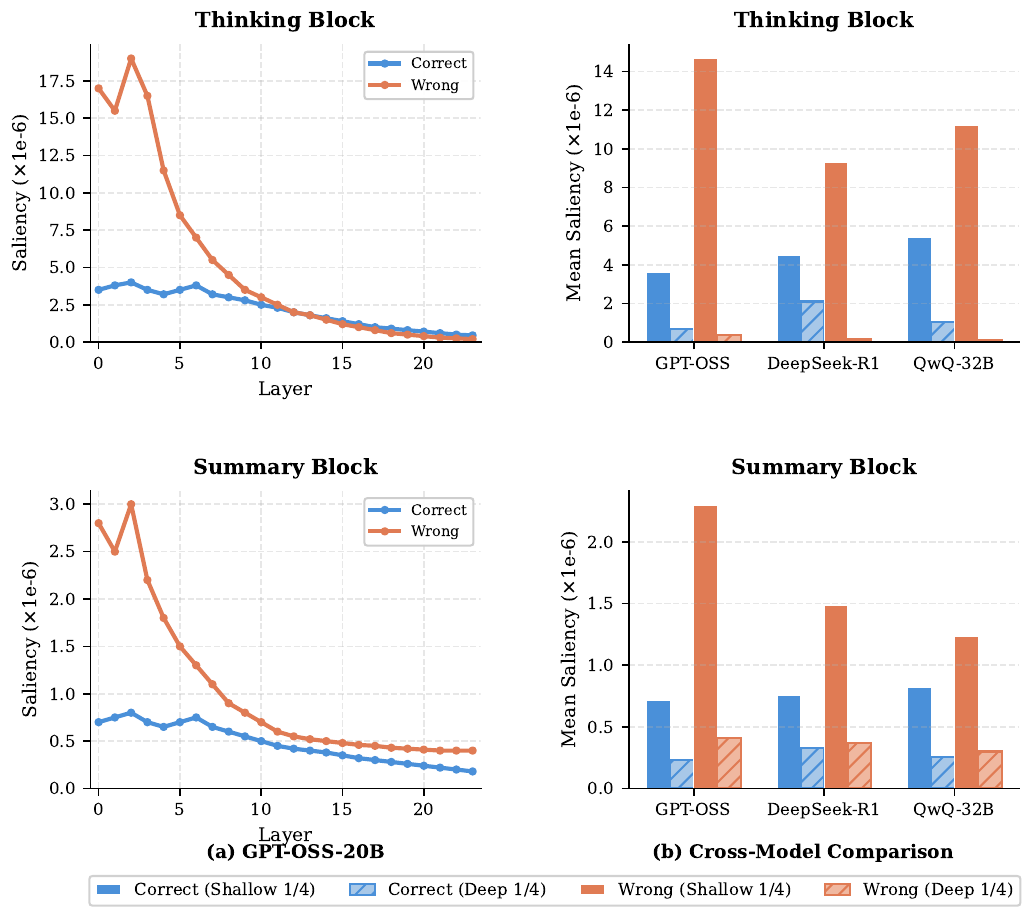}
  \caption{
  \textbf{Layer-wise saliency intensities across three models.}
  (a) GPT-OSS-20B: thinking and summary self-intensities for correct vs.\ error traces.
  (b) R1-Distill-32B and QwQ-32B show the same pattern: stronger shallow lock-in and summary self-reinforcement in error traces.
  }
  \label{fig:cross-model-saliency}
\end{figure}

Figure~\ref{fig:cross-model-saliency} shows that, for error traces,
thinking self-intensity is higher in shallow layers and decreases faster with depth, 
while summary self-saliency rises earlier and becomes more self-focused in deep layers, 
with much weaker connections back to earlier thinking steps than in correct traces.

Across models, error traces show higher shallow thinking self-intensity and earlier/higher summary self-intensity, consistent with Shallow Lock-in and Deep Decay.

\section{Method}
\label{sec:method}

Guided by the Step-Saliency patterns in \S\ref{sec:metrics}, we design \textbf{StepFlow}, a test-time intervention with two components: \textbf{Odds-Equal Bridge (OEB)} for shallow layers and \textbf{Step Momentum Injection (SMI)} for deep layers.

\subsection{Odds-Equal Bridge (OEB)}

\label{sec:oeb}

OEB aims to avoid a situation where almost all influence mass sits on the
current thinking step and its neighbours, while earlier context is ignored.

\paragraph{Group-wise proxy target.}
For a fixed query position $t$ in one head and one layer, let $p_t$ be the causal attention distribution over past tokens for this head.
Step-Saliency uses attention--gradient products for \emph{offline}
diagnosis, while OEB uses $p_t$ as a lightweight proxy during decoding
to enforce a minimum mass on the bridge region.
Using the segmentation from Step-Saliency, we split the keys into three
disjoint sets: the current segment $\mathcal{S}$, a \emph{bridge} segment
$\mathcal{B}$ that represents earlier context we want to preserve (e.g.,
the question while generating analysis, or the analysis while generating the summary), and all remaining tokens $\mathcal{O}$.
We define the current group masses
\begin{equation}
p_t(g) = \sum_{k \in g} p_t(k), \qquad
g \in \{\mathcal{S}, \mathcal{B}, \mathcal{O}\}.
\label{eq:oeb-group-mass}
\end{equation}
We keep $p_t(\mathcal{O})$ fixed and set a soft lower bound on the
bridge mass:
\begin{equation}
\begin{aligned}
\tau_{\mathcal{B}}
&= \min\!\left(
    \sqrt{\frac{|\mathcal{B}|}{|\mathcal{B}| + |\mathcal{S}|}},
    \,\tau_{\max}
  \right), \\
\tau_{\mathcal{S}}
&= 1 - p_t(\mathcal{O}) - \tau_{\mathcal{B}}.
\end{aligned}
\label{eq:oeb-target-mass}
\end{equation}
We apply OEB only when the bridge mass falls below the bound,
$p_t(\mathcal{B}) < \tau_{\mathcal{B}}$; otherwise we leave the logits
unchanged.
This schedule keeps the bridge on the same order of magnitude as the
current segment instead of letting its mass shrink to nearly zero.
Intuitively, the lower bound grows with the relative size of the bridge region, the square-root dampens extreme length effects, and $\tau_{\max}$ caps the intervention so OEB cannot dominate the attention distribution.
A single scalar $\tau_{\max}$ is chosen per model; we show in Appendix~\ref{app:hparam} that accuracy is robust across a wide range of $\tau_{\max}$ values.

\paragraph{KL projection on logits.}
Let $\mathbf{z}$ denote the attention logits (before the causal-mask softmax) for query position $t$ in a given layer/head, so that $p_t = \mathrm{softmax}(\mathbf{z})$.
When $p_t(\mathcal{B}) < \tau_{\mathcal{B}}$, we seek a new distribution
$q_t$ that stays as close as possible to $p_t$ in KL divergence while
enforcing $q_t(\mathcal{O}) = p_t(\mathcal{O})$ and
$q_t(\mathcal{B}) = \tau_{\mathcal{B}}$, and hence
$q_t(\mathcal{S}) = \tau_{\mathcal{S}}$.
This gives a small projection problem
$\arg\min_{q_t} \mathrm{KL}(q_t \,\|\, p_t)$ under linear constraints on
group totals,
which can be viewed as a constrained Bregman (KL) projection~\cite{banerjee2005clustering}.
Under the softmax parameterization, the solution reduces to a simple
group-wise shift of the scores:
\begin{equation}
z'_k \;=\; z_k + \lambda_g,
\qquad
k \in g,\; g \in \{\mathcal{S}, \mathcal{B}\},
\label{eq:oeb-shift}
\end{equation}
where, when the constraint is active,
\begin{equation}
\lambda_{\mathcal{B}} = \log\frac{\tau_{\mathcal{B}}}{p_t(\mathcal{B})}
\quad\text{and}\quad
\lambda_{\mathcal{S}} = \log\frac{\tau_{\mathcal{S}}}{p_t(\mathcal{S})}.
\label{eq:oeb-lambdas}
\end{equation}
Scores in $\mathcal{O}$ are left unchanged.
The group-wise shift in Eq.~\eqref{eq:oeb-shift} ensures that when bridge mass falls below the threshold, attention logits are adjusted to maintain a minimum share on the bridge region, preventing the collapse of influence onto the current segment alone.

\subsection{Step Momentum Injection (SMI)}
\label{sec:smi}

SMI targets deep-layer decay.
As depth increases, thinking saliency tends to drift toward background, especially on error traces.
SMI treats deep layers as a leaky integrator and nudges them towards a mild accumulator: it preserves a small amount of step-level content while keeping the backbone computation intact.
\paragraph{Step segmentation.}
We reuse the step segmentation defined for Step-Saliency: the thinking segment is split into steps $\Gamma_1,\dots,\Gamma_K$, and the summary tokens form a summary segment $\Gamma_{K+1}$.
Boundaries are detected using the model’s analysis marker plus simple punctuation and template cues, and StepFlow is empirically robust to small boundary shifts (Appendix~\ref{app:seg-robust}).
\begin{table*}[t]
\centering
\caption{Accuracy (\%) on six benchmarks (columns) across multiple backbones.}
\label{tab:reasoning-coding-main}
\small
\setlength{\tabcolsep}{5pt}
\renewcommand{\arraystretch}{0.90}
\begin{tabular}{l *{6}{r}}
    \toprule
    \textbf{Method} &
    \multicolumn{5}{c}{\textbf{Math / Science}} &
    \textbf{Code} \\
    &
    AIME24 &
    AIME25 &
    AMC23 &
    MATH-500 &
    GPQA-D &
    LiveCodeBench \\
    \midrule
    DeepSeek-R1-Distill-Qwen 7B   & 54.0 & 39.2 & 82.5 & 92.8 & 49.1 & 37.6 \\
    \quad+ Budget Forcing (S1)    & 56.2 & 39.5 & 84.0 & 93.1 & 51.3 & 39.8 \\
    \quad+ Hint-Infer (Round1)             & 57.5 & 43.3 & 85.2 & 93.0 & 56.6 & 45.2 \\
    \quad+ Plan-and-Solve (PS+)            & 58.8 & 42.6 & 86.0 & 93.3 & 52.8 & 42.0 \\
    \quad+ Attn-Interv.            & 56.5 & 41.0 & 84.5 & 93.0 & 51.8 & 40.5 \\
    \quad+ Act.\ Steering          & 57.0 & 41.5 & 84.8 & 93.1 & 52.0 & 41.2 \\
    \rowcolor{gray!10}
    \quad+ \textbf{StepFlow}               & \textbf{62.5} & \textbf{43.8} & \textbf{88.0} & \textbf{93.8} & \textbf{57.6} & \textbf{47.1} \\
    \addlinespace[0.08em]
    DeepSeek-R1-Distill-Qwen 14B  & 69.7 & 50.2 & 87.1 & 93.9 & 59.1 & 53.1 \\
    \quad+ Budget Forcing (S1)    & 70.5 & 54.8 & 90.0 & 94.1 & 60.2 & 54.8 \\
    \quad+ Hint-Infer (Round1)             & 71.2 & 53.0 & 88.5 & 94.1 & 61.0 & 56.2 \\
    \quad+ Plan-and-Solve (PS+)            & 71.8 & 54.5 & 89.2 & 94.1 & 61.8 & 57.0 \\
    \quad+ Attn-Interv.            & 70.8 & 52.5 & 88.0 & 94.0 & 60.5 & 55.5 \\
    \quad+ Act.\ Steering          & 71.0 & 53.2 & 88.2 & 94.0 & 60.8 & 55.0 \\
    \rowcolor{gray!10}
    \quad+ \textbf{StepFlow}               & \textbf{72.1} & \textbf{57.7} & \textbf{89.7} & \textbf{94.2} & \textbf{63.1} & \textbf{59.9} \\
    \addlinespace[0.08em]
    DeepSeek-R1-Distill-Qwen 32B  & 72.6 & 54.9 & 93.8 & 94.3 & 62.1 & 57.2 \\
    \quad+ Budget Forcing (S1)    & 73.2 & 56.5 & 94.9 & 94.9 & 63.8 & 58.5 \\
    \quad+ Hint-Infer (Round1)             & 73.0 & 57.8 & 94.5 & 94.8 & 63.5 & 59.8 \\
    \quad+ Plan-and-Solve (PS+)            & 74.0 & 59.5 & 94.9 & 95.0 & 63.8 & 60.5 \\
    \quad+ Attn-Interv.            & 73.5 & 57.5 & 94.2 & 94.5 & 62.9 & 59.8 \\
    \quad+ Act.\ Steering          & 74.0 & 58.2 & 94.5 & 94.6 & 64.0 & 58.5 \\
    \rowcolor{gray!10}
    \quad+ \textbf{StepFlow}               & \textbf{74.5} & \textbf{66.7} & \textbf{95.3} & \textbf{95.6} & \textbf{64.5} & \textbf{63.0} \\
    \midrule
    GPT-OSS-20B low               & 41.8 & 39.1 & 83.9 & 86.2 & 57.1 & 52.4 \\
    \quad+ Budget Forcing (S1)    & 44.9 & 40.5 & 85.2 & 87.0 & 58.5 & 53.8 \\
    \quad+ Hint-Infer (Round1)             & 44.5 & 41.8 & 86.5 & 86.8 & 59.8 & 56.2 \\
    \quad+ Plan-and-Solve (PS+)            & 45.6 & 42.8 & 87.2 & 87.5 & 60.8 & 57.0 \\
    \quad+ Attn-Interv.            & 43.5 & 40.8 & 85.5 & 86.8 & 58.8 & 55.0 \\
    \quad+ Act.\ Steering          & 44.0 & 41.2 & 85.8 & 87.0 & 59.2 & 55.5 \\
    \rowcolor{gray!10}
    \quad+ \textbf{StepFlow}               & \textbf{47.9} & \textbf{46.5} & \textbf{90.0} & \textbf{89.6} & \textbf{64.0} & \textbf{62.3} \\
    \addlinespace[0.08em]
    GPT-OSS-20B medium            & 63.4 & 62.0 & 94.2 & 89.2 & 65.2 & 70.0 \\
    \quad+ Budget Forcing (S1)    & 65.2 & 63.5 & 94.6 & 89.6 & 66.5 & 72.5 \\
    \quad+ Hint-Infer (Round1)             & 65.0 & 64.8 & 94.8 & 89.5 & 67.8 & 74.5 \\
    \quad+ Plan-and-Solve (PS+)            & 65.6 & 66.0 & 95.0 & 89.9 & 68.5 & 75.8 \\
    \quad+ Attn-Interv.            & 64.8 & 66.2 & 94.5 & 89.5 & 66.0 & 75.8 \\
    \quad+ Act.\ Steering          & 65.5 & 64.5 & 94.6 & 89.6 & 66.8 & 72.5 \\
    \rowcolor{gray!10}
    \quad+ \textbf{StepFlow}               & \textbf{66.0} & \textbf{69.2} & \textbf{95.5} & \textbf{90.5} & \textbf{70.3} & \textbf{79.5} \\
    \midrule
    QwQ-32B-Preview               & 50.0 & 40.0 & 80.0 & 90.6 & 58.1 & 41.4 \\
    \quad+ Budget Forcing (S1)    & 54.5 & 41.5 & 82.0 & 91.0 & 59.5 & 46.2 \\
    \quad+ Hint-Infer (Round1)             & 52.8 & 42.8 & 83.5 & 90.3 & 60.8 & 44.8 \\
    \quad+ Plan-and-Solve (PS+)            & 53.8 & 44.0 & 84.5 & 91.2 & 61.5 & 46.5 \\
    \quad+ Attn-Interv.            & 52.0 & 42.2 & 82.5 & 90.8 & 59.8 & 44.5 \\
    \quad+ Act.\ Steering          & 53.0 & 43.5 & 83.0 & 91.0 & 59.2 & 45.0 \\
    \rowcolor{gray!10}
    \quad+ \textbf{StepFlow}               & \textbf{57.3} & \textbf{48.0} & \textbf{91.0} & \textbf{92.0} & \textbf{63.3} & \textbf{50.3} \\
    \bottomrule
\end{tabular}
\end{table*}

\paragraph{Step-level residual link.}
At the boundary between $\Gamma_i$ and $\Gamma_{i+1}$, we build a step-level momentum vector from the deep-layer value states of step $\Gamma_i$.
In practice, $\mathbf{v}_k$ is taken as the multi-head attention value projection output (concatenated over heads) at the selected deep layers.
Let $\{\mathbf{v}_k\}_{k \in \Gamma_i}$ be the corresponding value vectors and define a single step summary
\begin{equation}
\mathbf{m}_{\text{prev}} = \frac{1}{|\Gamma_i|} \sum_{k \in \Gamma_i} \mathbf{v}_k.
\label{eq:smi-mprev}
\end{equation}
We then inject this momentum as a residual into the hidden state of the first token $t$ of $\Gamma_{i+1}$ in a subset of deep layers:
\begin{equation}
\mathbf{h}'_t = \mathbf{h}_t + \alpha\,\mathbf{m}_{\text{prev}}.
\label{eq:smi-inject}
\end{equation}
We add the residual on the layer's residual stream before the MLP block.
The scalar $\alpha$ is a small, fixed coefficient per model; details and implementation variants are given in Appendix~\ref{app:hparam}.
By injecting the momentum vector at step boundaries, SMI maintains a connection from earlier thinking steps to later positions, counteracting the tendency for deep layers to lose saliency on the thinking segment.

\begin{algorithm}[t]
\caption{\textsc{StepFlow} single-pass decoding}
\label{alg:stepflow}
\begin{algorithmic}[1]
\small
\Require $\mathcal{M},\,\mathcal{L}_{\mathrm{sh}},\,\mathcal{L}_{\mathrm{dp}},\,\tau_{\max},\,\alpha$
\For{$t=1,2,\dots$}
  \State $(\mathcal{S},\mathcal{B},\mathcal{O}) \gets \textsc{PartitionKeys}(t)$
  \State $\tau_{\mathcal{B}}\gets \min\!\Big(\sqrt{\frac{|\mathcal{B}|}{|\mathcal{B}|+|\mathcal{S}|}},\,\tau_{\max}\Big)$
  \For{$\ell \in \mathcal{L}_{\mathrm{sh}}$}
      \State $p_g \gets \sum_{k\in g}\mathrm{softmax}(\mathbf{z}^{(\ell)}_t)[k]\ (g\in\{\mathcal{S},\mathcal{B},\mathcal{O}\})$
      \If{$p_{\mathcal{B}} < \tau_{\mathcal{B}}$}
  \State $\tau_{\mathcal{S}}\gets 1-p_{\mathcal{O}}-\tau_{\mathcal{B}}$
  \State $\lambda_{\mathcal{B}}\gets \log\frac{\tau_{\mathcal{B}}}{p_{\mathcal{B}}},\ 
         \lambda_{\mathcal{S}}\gets \log\frac{\tau_{\mathcal{S}}}{p_{\mathcal{S}}}$
  \State $\mathbf{z}^{(\ell)}_t[\mathcal{B}] \mathrel{+}= \lambda_{\mathcal{B}};\ 
         \mathbf{z}^{(\ell)}_t[\mathcal{S}] \mathrel{+}= \lambda_{\mathcal{S}}$
\EndIf
  \EndFor
  \If{\textsc{IsStepBoundary}$(t)$}
    \State $\mathbf{m}_{\mathrm{prev}}\gets \frac{1}{|\Gamma_i|}\sum_{k\in\Gamma_i}\mathbf{v}_k$
    \For{$\ell\in\mathcal{L}_{\mathrm{dp}}$}
      \State $\mathbf{h}^{(\ell)}_{t} \mathrel{+}= \alpha\,\mathbf{m}_{\mathrm{prev}}$
    \EndFor
  \EndIf
  \State $x_{t+1}\sim p_{\mathcal{M}}(\cdot\mid x_{\le t})$
  \If{$x_{t+1}=\textsc{EoS}$} \State \textbf{break} \EndIf
\EndFor
\end{algorithmic}
\end{algorithm}

\section{Experiment}

\subsection{Experimental Setup}
\noindent\textbf{Model.}
We focus on open-weight large reasoning models that emit explicit chain-of-thought.
Our main backbones are DeepSeek-R1-Distill-Qwen (7B/14B/32B)~\cite{guo2025deepseek}, GPT-OSS-20B~\cite{agarwal2025gpt}, and QwQ-32B-Preview~\cite{team2025qwq}.

\noindent\textbf{Evaluation.}
We evaluate on six challenging benchmarks: AIME24, AIME25, AMC23, MATH-500~\cite{hendrycks2021measuring}, GPQA-Diamond~\cite{rein2024gpqa}, and LiveCodeBench~\cite{jain2024livecodebench}.
They are widely used to test multi-step reasoning.
We use the same decoding setup for all models and report accuracy (Appendix~\ref{app:experiment-settings}).
To make the results more stable under random sampling, we average over 16 sampled solutions per problem for AIME24/25 and AMC23, and 8 for GPQA-Diamond; MATH-500 and LiveCodeBench use the standard single-sample setting.
All methods are applied to each sample in one pass (no multi-pass voting).
All baselines and StepFlow share identical decoding hyperparameters, stop conditions, and answer extraction rules (Appendix~\ref{app:experiment-settings}).

\noindent\textbf{Baselines.}
We compare StepFlow against prompt-only baselines (Plan-and-Solve (PS+)~\cite{wang2023plan} and Hint-Infer (Round1)~\cite{li2025start}), decode-level baselines (Budget Forcing (S1)~\cite{muennighoff2025s1}), internal intervention baselines (Attn-Interv.~\cite{yan2025don} and Activation Steering~\cite{zhang2023tell}), and the unmodified backbone; baseline definitions and settings are summarized in Appendix~\ref{app:baselines}.

\subsection{Main Results on Reasoning Benchmarks}

Prior test-time methods for reasoning models either extend generation length~\cite{muennighoff2025s1} or require multiple forward passes for voting~\cite{wang2023self}.
StepFlow instead repairs information flow \emph{inside} an LRM within a single decoding run, without multi-pass voting, with moderate overhead (Appendix~\ref{app:decode}).

Table~\ref{tab:reasoning-coding-main} shows that StepFlow consistently improves all six backbones across math, science, and code benchmarks.
Gains are most pronounced on competition-style problems that demand long reasoning chains. For example, StepFlow adds +11.8 points on AIME25 for R1-Distill-32B (54.9$\rightarrow$66.7), and also improves LiveCodeBench on GPT-OSS-20B medium by +9.5 (70.0$\rightarrow$79.5).
This is consistent with our diagnosis: repairing shallow lock-in and deep decay matters most when reasoning requires information to be propagated across many intermediate steps.
To understand which error types benefit, we manually categorized all 60 AIME 24/25 problems where baselines failed but StepFlow succeeded. Arithmetic carry-forward (34\%) and premise forgetting (38\%) together account for 72\% of corrections---precisely the categories where cross-step propagation matters. Conceptual errors (10--14\%) are rarely fixed, confirming that StepFlow repairs information flow rather than knowledge (full taxonomy in Appendix~\ref{app:error-taxonomy}).

\subsection{Effect on Information Flow}

\begin{figure}[t]
  \centering
  \includegraphics[width=\linewidth]{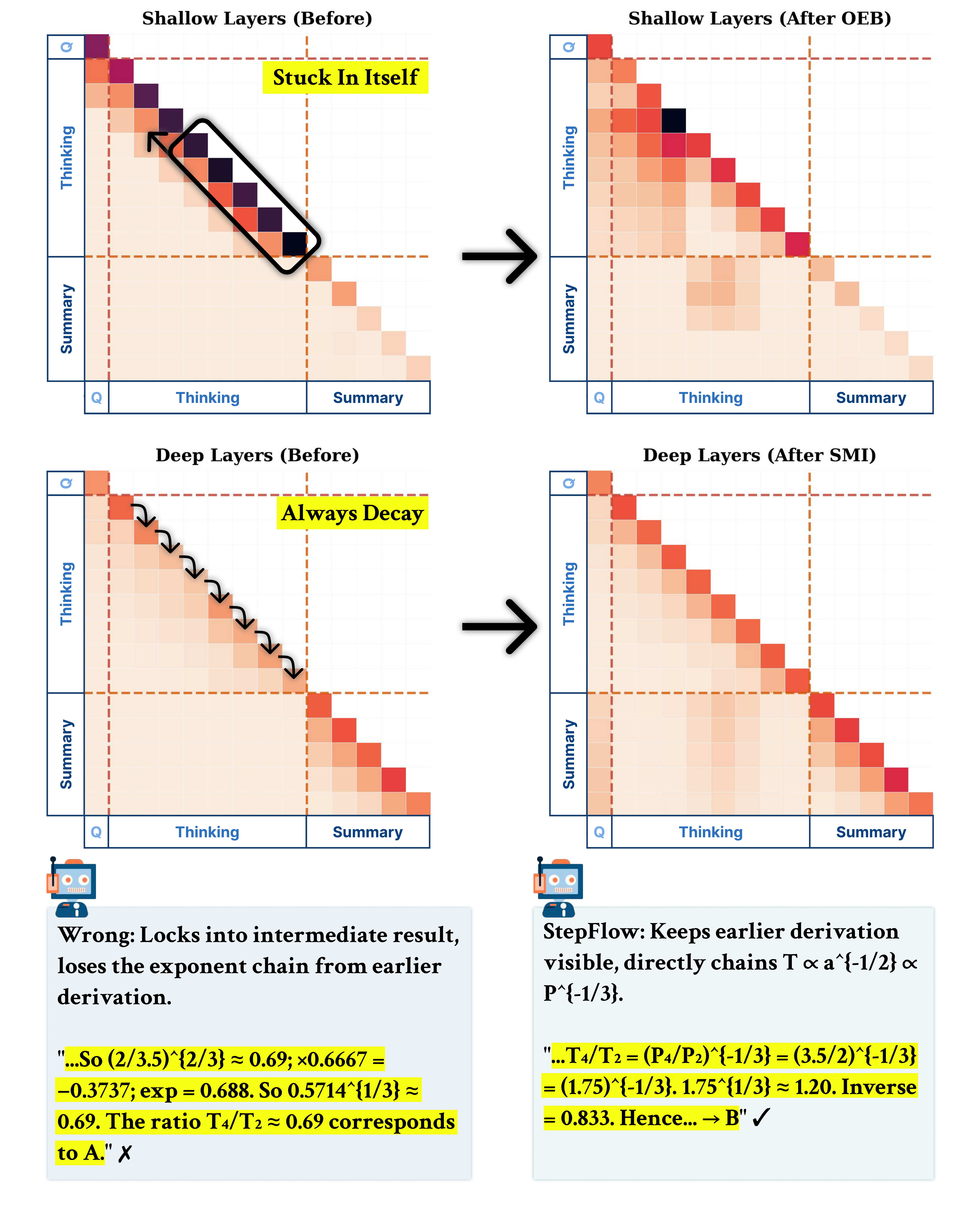}
  \caption{
  Effect of StepFlow on Step-Saliency (representative error trace).
  See Appendix~\ref{app:case-studies} for the full response.
  }
  \label{fig:stepflow-effect}
\end{figure}

In the Step-Saliency view, each layer induces a step-to-step transition matrix over the question, thinking steps, and summary. Figure~\ref{fig:stepflow-effect} shows a representative error trace: in shallow layers, OEB reduces near-diagonal self-loops and strengthens the bridge region; in deep layers, SMI restores broader links from the summary into the thinking segment. See Appendix~\ref{app:case-studies} for a full case study with model outputs.

\begin{figure}[t]
  \centering
  \includegraphics[width=\linewidth]{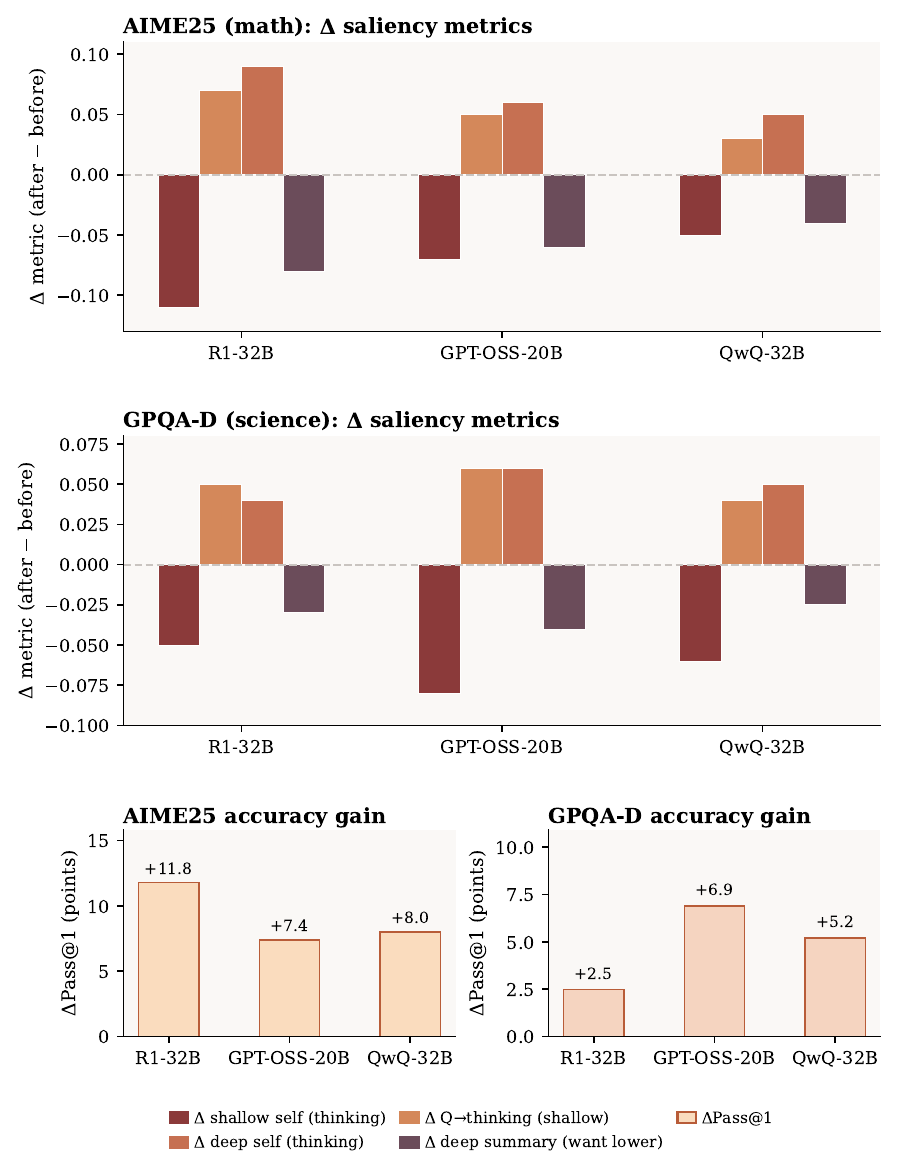}
\caption{
Effect of StepFlow on AIME25 and GPQA-D across three backbones.
Top: $\Delta$ Step-Saliency ($I_T$, $I_S$; \S\ref{sec:metrics}) over error cases. Bottom: Accuracy gain.
}
  \label{fig:aime25-metric}
\end{figure}

Beyond a single trace, we also aggregate Step-Saliency statistics over all AIME25 error cases (Figure~\ref{fig:aime25-metric}).
For each backbone, StepFlow consistently reduces shallow thinking self-intensity and deep summary self-intensity, while increasing shallow question$\rightarrow$thinking mass.
These shifts mean that shallow layers rely less on the just-written step and use the question more, and deep layers keep a stronger link from the thinking segment into the summary.

\subsection{Ablation Analysis}

\begin{table}[t]
\centering
\footnotesize
\setlength{\tabcolsep}{1pt}
\caption{Ablation of StepFlow components on GPT-OSS-20B medium (Accuracy, \%).}
\label{tab:stepflow-ablate}
\begin{tabular}{lccc}
\toprule
Method                  & AIME25 & GPQA\textsubscript{D} & LiveCodeBench \\
\midrule
Baseline                & 62.0   & 65.2   & 70.0 \\
+ OEB only (shallow)    & 64.5   & 66.7   & 74.5 \\
+ SMI only (deep)       & 64.0   & 67.2   & 75.0 \\
+ OEB + SMI (StepFlow)  & 69.2   & 70.3   & 79.5 \\
\bottomrule
\end{tabular}
\end{table}

\begin{table}[t]
\centering
\small
\caption{Accuracy (\%) on LiveCodeBench by difficulty for GPT-OSS-20B medium with and without StepFlow.}
\label{tab:lcb-difficulty-gptoss}
\begin{tabular}{lccc}
\toprule
Model & Easy & Medium & Hard \\
\midrule
GPT-OSS-20B        & 90.1 & 70.2 & 37.8 \\
+StepFlow          & \textbf{93.5} & \textbf{84.0} & \textbf{52.0} \\
\bottomrule
\end{tabular}
\end{table}

We separate the two components of StepFlow on GPT-OSS-20B in Table~\ref{tab:stepflow-ablate}.
Using only OEB already helps on all three benchmarks, and only SMI helps a bit more, especially on GPQA\textsubscript{D} and LiveCodeBench; combining both gives the largest gains.
This matches the diagnostics: shallow layers need less lock-in on the current step, and deep layers need a stronger link from the thinking segment when the summary is produced.

LiveCodeBench is broken into easy, medium, and hard problems in Table~\ref{tab:lcb-difficulty-gptoss}.
Easy items are close to a ceiling and StepFlow adds only a few extra solved cases, but the larger jumps on medium (+13.8 points) and hard (+14.2 points) items suggest that StepFlow is most helpful when the model has to carry a longer and more fragile reasoning chain.

\subsection{Layer Coverage and Decoding Cost}

\begin{table}[!t]
\centering
\footnotesize
\caption{Effect of StepFlow layer coverage (OEB on bottom-$k$ layers, SMI on top-$k$ layers) on three LRMs (Accuracy, \%).}
\label{tab:layer-sensitivity}
\resizebox{\columnwidth}{!}{%
\begin{tabular}{lccc}
\toprule
\multicolumn{4}{c}{\textbf{AIME25}} \\
\midrule
Active layers (OEB / SMI) & R1-Distill-32B & GPT-OSS-20B & QwQ-32B \\
\midrule
Baseline (no StepFlow)     & 54.9 & 39.1 & 40.0 \\
Bottom / top 1/4 of layers & 66.7 & 46.5 & 48.0 \\
Bottom / top 1/3 of layers & 64.0 & 44.0 & 45.2 \\
Bottom / top 1/2 of layers & 62.3 & 42.0 & 43.0 \\
\midrule
\multicolumn{4}{c}{\textbf{GPQA\textsubscript{D}}} \\
\midrule
Active layers (OEB / SMI) & R1-Distill-32B & GPT-OSS-20B & QwQ-32B \\
\midrule
Baseline (no StepFlow)     & 62.1 & 57.1 & 58.1 \\
Bottom / top 1/4 of layers & 64.5 & 64.0 & 63.3 \\
Bottom / top 1/3 of layers & 63.2 & 62.0 & 61.2 \\
Bottom / top 1/2 of layers & 62.4 & 60.5 & 60.0 \\
\bottomrule
\end{tabular}}
\end{table}

Table~\ref{tab:layer-sensitivity} shows the best gains when activating only the bottom/top quarter of layers; broader coverage (one third or one half) slightly reduces gains but remains above the baseline, consistent with Step-Saliency indicating most leverage in shallow/deep bands.
Decoding efficiency (Table~\ref{tab:gpqa-decode-time}) and uncertainty estimates are reported in Appendix~\ref{app:decode} and Appendix~\ref{app:uncertainty}.
\subsection{Compute-Normalized Comparison}
\label{sec:pareto}

To assess whether StepFlow's gains justify its overhead, we compare it against longer generation and self-consistency (SC) at matched compute budgets.
Table~\ref{tab:pareto} reports results on AIME 24/25 (averaged) for R1-Distill-32B; full results for all three backbones are in Appendix~\ref{app:pareto-full}.

\begin{table}[t]
\centering
\small
\setlength{\tabcolsep}{4pt}
\caption{Compute-normalized comparison on AIME 24/25 (averaged accuracy, \%) for R1-Distill-32B.}
\label{tab:pareto}
\begin{tabular}{lcc}
\toprule
Method & Compute & Acc.\ (\%) \\
\midrule
Baseline        & 1.0$\times$ & 63.8 \\
Longer gen      & ${\sim}$1.35$\times$ & 65.0 \\
StepFlow        & ${\sim}$1.35$\times$ & \textbf{70.6} \\
SC ($k$=2)      & ${\sim}$2.0$\times$ & 66.5 \\
StepFlow+SC ($k$=2) & ${\sim}$2.7$\times$ & \textbf{73.5} \\
SC ($k$=4)      & ${\sim}$4.0$\times$ & 68.5 \\
SC ($k$=8)      & ${\sim}$8.0$\times$ & 70.2 \\
\bottomrule
\end{tabular}
\end{table}

At matched compute (${\sim}$1.35$\times$), StepFlow achieves 5.7$\times$ the gain of longer generation.
Matching StepFlow's accuracy requires SC with $k$$\approx$8 at 8.0$\times$ compute.
StepFlow is also composable: StepFlow+SC($k$=2) at ${\sim}$2.7$\times$ outperforms SC($k$=4) at 4.0$\times$, showing that StepFlow and sampling address orthogonal failure modes.

\section{Discussion}

\paragraph{Causal status of the diagnostic.}
StepFlow's gains exhibit three forms of specificity---complementary OEB/SMI profiles across benchmarks (Table~\ref{tab:stepflow-ablate}), optimal performance only at the predicted layer bands (Table~\ref{tab:layer-sensitivity}), and selective correction of propagation errors at 5--7$\times$ the rate of conceptual errors (Appendix~\ref{app:error-taxonomy})---which rule out a generic regularization account. Nevertheless, the causal link between the diagnosed saliency patterns and the observed improvements remains suggestive rather than formally proved.

\paragraph{Memory versus reasoning.}
StepFlow corrects cases where a correct intermediate result exists but fails to propagate to later steps. Whether such failures are best characterized as ``memory'' or ``reasoning'' errors is an open question; recent attempts to disentangle the two~\cite{jin2025disentangling,li2025demystifying,wu2025knowledge} all rely on external annotations rather than model-internal definitions. We view active retrieval failure during generation as functionally relevant to reasoning, while acknowledging this distinction warrants further investigation.

\section{Conclusion}

We introduce Step-Saliency, a step-level diagnostic that aggregates token saliency into question--thinking--summary maps and reveals two depth-wise failure modes in LRMs (Shallow Lock-in and Deep Decay). Guided by these patterns, we propose StepFlow, a lightweight test-time intervention that repairs information flow and consistently improves accuracy on math, science, and coding benchmarks across multiple LRMs without retraining.

\section*{Limitations}

\paragraph{Model-specific calibration.}
StepFlow assumes a shallow/deep split that depends on the backbone and is chosen on a small held-out split.
Our layer-coverage study suggests a broad sweet spot, but we do not provide a fully automatic way to pick these ranges.

\paragraph{Intervention design space.}
StepFlow is one concrete, saliency-inspired intervention.
We do not explore other designs (e.g., head-level steering or value-space projections), although they could be analyzed in the same framework.

\paragraph{Error-type coverage.}
Our error taxonomy (Appendix~\ref{app:error-taxonomy}) shows that StepFlow primarily corrects information-propagation errors (72\%) rather than conceptual ones (10--14\%). A finer-grained breakdown within each propagation category and extension to benchmarks beyond AIME are left for future work.

\bibliography{reference}

\appendix

\section{Step-Saliency Analysis}
\label{app:saliency}

\subsection{Computation Protocol}
\label{app:saliency-computation}

This subsection clarifies the aggregation protocol in Eqs.~(1)--(3) of the main text.

\paragraph{Dependence on $t$.}
The influence matrix $\mathbf{I}^{(\ell)}$ in Eq.~(1) is computed row by row: for each query position $t$, we backpropagate through the token-level loss $\mathcal{L}_t$ to obtain gradients with respect to the $t$-th row of attention weights.
Due to causal masking, only entries $(t,k)$ with $k \le t$ are non-zero.
The full $T \times T$ matrix is obtained by stacking all rows.

\paragraph{Implementation constants.}
In Eq.~(2), we use $\varepsilon = 10^{-8}$ to prevent division by zero during row normalization.

\paragraph{Depth aggregation.}
For the depth-collapsed Step-Saliency maps shown in figures, we average $M^{(\ell)}$ over all $L$ layers:
$\bar{M}_{j \leftarrow i} = \frac{1}{L} \sum_{\ell=1}^{L} M^{(\ell)}_{j \leftarrow i}$.
For layer-wise analysis (shallow vs.\ deep), we average over the bottom 1/4 or top 1/4 of layers, respectively.

\subsection{Segmentation Robustness}
\label{app:seg-robust}

\paragraph{Boundary detection.}
We detect step boundaries using the model's analysis marker (when available), plus simple formatting cues such as sentence-ending punctuation and newline characters.
Pure-digit strings and table/separator lines are filtered out.

\paragraph{Empirical accuracy of boundary detection.}
We manually inspected 100 randomly sampled traces from GPQA-D and AIME24.
The heuristic detector correctly identified 99\% of step boundaries.
The few failures occur when the model produces repeated delimiters (e.g., multiple consecutive newlines), which is rare in typical LRM outputs.

\paragraph{Robustness to boundary perturbations.}
To stress-test the method, we artificially perturb boundaries:
(i) \emph{shift}---uniform offset by $\pm k$ tokens,
(ii) \emph{dropout}---randomly removing 25\% of boundaries,
(iii) \emph{insertion}---randomly adding 25\% spurious boundaries,
(iv) \emph{combined}---simultaneous 50\% dropout and 50\% insertion, and
(v) \emph{random uniform}---replacing all detected boundaries with uniformly random positions.
Table~\ref{tab:seg-robust} shows that even under extreme perturbations, accuracy degrades by at most 2.0 points from the default, and StepFlow consistently outperforms the baseline.
Since manual inspection shows 1\% detection error in practice, the method operates well within its robustness margin.

\begin{table}[t]
\centering
\small
\caption{Effect of boundary perturbations on GPQA-D accuracy (\%).}
\label{tab:seg-robust}
\setlength{\tabcolsep}{3pt}
\begin{tabular}{llccc}
\toprule
Perturbation & Level & R1-32B & GPT-OSS & QwQ-32B \\
\midrule
Shift        & $\pm 1$ tok  & 64.2 & 69.8 & 63.0 \\
             & $\pm 3$ tok  & 64.0 & 69.5 & 62.7 \\
\midrule
Dropout      & 25\%  & 63.3 & 69.2 & 62.1 \\
Insertion    & 25\%  & 63.1 & 68.9 & 62.0 \\
Dropout+Ins  & 50\% each & 62.8 & 67.5 & 60.5 \\
Random unif  & ---        & 62.5 & 66.8 & 59.5 \\
\midrule
\multicolumn{2}{l}{Default}    & 64.5 & 70.3 & 63.3 \\
\multicolumn{2}{l}{No StepFlow} & 62.1 & 65.2 & 58.1 \\
\bottomrule
\end{tabular}
\end{table}

\section{StepFlow Implementation}
\label{app:stepflow-impl}

\subsection{Hyperparameter Selection}
\label{app:hparam}

\begin{figure}[t]
  \centering
  \includegraphics[width=\columnwidth]{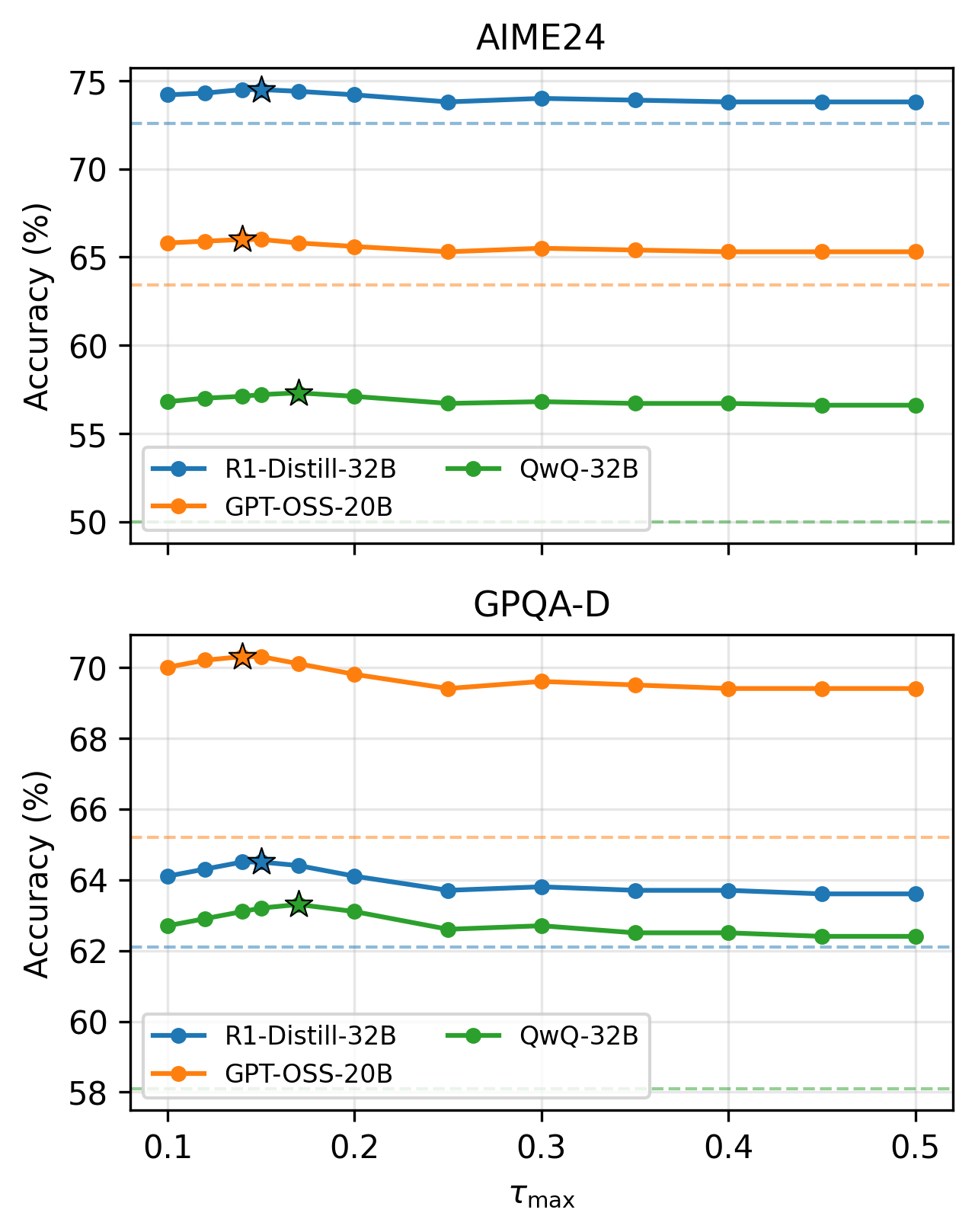}
  \caption{Accuracy vs.\ $\tau_{\max}$ on AIME24 and GPQA-D. Dashed lines show baseline accuracy; stars mark the chosen $\tau^\star$.}
  \label{fig:taumax-sensitivity}
\end{figure}

\paragraph{OEB threshold $\tau_{\max}$.}
Figure~\ref{fig:taumax-sensitivity} shows accuracy as a function of $\tau_{\max}$ on AIME24 and GPQA-D.
Across all three backbones, accuracy consistently improves over the baseline for any $\tau_{\max} \in [0.1, 0.5]$; the curves are nearly flat, with peak-to-trough variation under 1 point.
This indicates that the exact choice of $\tau_{\max}$ has negligible impact on performance, and a single default value (e.g., $\tau_{\max} = 0.15$) works well across all models without any tuning.

\paragraph{SMI residual scale $\alpha$.}
The only free hyperparameter in SMI is the residual scale $\alpha$.
For each backbone, we select $\alpha$ once and reuse the same value for all benchmarks and all reported tables.

\paragraph{Grid and selection protocol.}
We sweep $\alpha$ over a small grid
$\{0.03,0.06,0.09\}$ and evaluate accuracy on two
representative benchmarks: AIME24 (competition-style math) and
GPQA-Diamond (graduate-level science).
For each model/benchmark pair, we record:
(i) the smallest scale $\alpha_{\mathrm{low}}$ that improves over the
no-intervention baseline,
(ii) the chosen scale $\alpha^\star$, and
(iii) the largest scale $\alpha_{\mathrm{high}}$ that does not reduce
accuracy by more than 1 point compared with $\alpha^\star$.
The last column in Table~\ref{tab:alpha-range} reports the accuracy range
achieved within this interval.

\paragraph{Observed robustness.}
Table~\ref{tab:alpha-range} shows that all three backbones admit a
non-trivial interval
$[\alpha_{\mathrm{low}}, \alpha_{\mathrm{high}}]$ on both AIME24 and
GPQA-D where SMI improves over the baseline.
Within each interval, the spread in accuracy is small (typically within
1--2 points), indicating that StepFlow does not rely on fine-tuning
$\alpha$.
In our main experiments, we simply fix $\alpha^\star$ per backbone
(based on the AIME24 sweep) and reuse it for all other benchmarks
without further tuning.

\begin{table}[t]
  \centering
  \small
  \caption{Sensitivity of the SMI residual scale $\alpha$ on AIME24 and
  GPQA-D. For each model/benchmark pair, 
  The last column reports the corresponding accuracy interval.}
  \label{tab:alpha-range}
  \resizebox{\columnwidth}{!}{%
  \begin{tabular}{lcccc}
    \toprule
    Model / Benchmark & $\alpha_{\text{low}}$ & $\alpha^\star$ & $\alpha_{\text{high}}$ & Range \\
    \midrule
    R1-Distill-32B / AIME24     & 0.03 & 0.06 & 0.09 & 74.9--75.5 \\
    R1-Distill-32B / GPQA-D     & 0.03 & 0.06 & 0.09 & 64.3--64.9 \\
    GPT-OSS / AIME24    & 0.03 & 0.06 & 0.09 & 65.4--66.0 \\
    GPT-OSS / GPQA-D    & 0.03 & 0.06 & 0.09 & 67.5--68.1 \\
    QwQ-32B / AIME24    & 0.03 & 0.06 & 0.09 & 56.9--57.5 \\
    QwQ-32B / GPQA-D    & 0.03 & 0.06 & 0.09 & 62.6--63.2 \\
    \bottomrule
  \end{tabular}%
  }
\end{table}

\subsection{Reproducibility Summary}
\label{app:repro}

We provide a compact reproducibility summary of the StepFlow configuration in Table~\ref{tab:stepflow-repro-summary}.

\paragraph{OEB group definition.}
To match the question--thinking--summary decomposition used by Step-Saliency, OEB uses the same coarse segmentation during decoding and defines group totals over keys for each attention head and query position $t$.
When $t$ is in the thinking segment, $\mathcal{S}$ are thinking tokens and $\mathcal{B}$ are question tokens; when $t$ is in the summary segment, $\mathcal{S}$ are summary tokens and $\mathcal{B}$ are thinking tokens.
All remaining keys are grouped into $\mathcal{O}$, and its total mass is kept fixed, matching Eq.~\eqref{eq:oeb-group-mass}--Eq.~\eqref{eq:oeb-lambdas}.

\paragraph{SMI value source and injection location.}
SMI triggers only at detected step boundaries inside the thinking segment.
At the first token of the next step, in selected deep layers, we compute a step momentum vector by mean-pooling the value states $\{\mathbf{v}_k\}$ from the previous step span, and inject it as a small residual into the hidden state used for the next step.
This corresponds to Eq.~\eqref{eq:smi-mprev} and Eq.~\eqref{eq:smi-inject} in the main text with a single scale $\alpha$.

\begin{table}[t]
\centering
\small
\caption{StepFlow reproducibility summary.}
\label{tab:stepflow-repro-summary}
\begin{tabular}{lcccc}
\toprule
Model & OEB layers & SMI layers & $\tau_{\max}$ & $\alpha$ \\
\midrule
R1-Distill-7B  & bottom 1/4 & top 1/4 & 0.15 & 0.06\\
R1-Distill-14B & bottom 1/4 & top 1/4 & 0.15 & 0.06\\
R1-Distill-32B & bottom 1/4 & top 1/4 & 0.15 & 0.06\\
GPT-OSS-20B    & bottom 1/4 & top 1/4 & 0.15 & 0.06\\
QwQ-32B        & bottom 1/4 & top 1/4 & 0.15 & 0.06\\
\bottomrule
\end{tabular}
\end{table}

\section{Experimental Details}
\label{app:experiments}

\subsection{Baselines}
\label{app:baselines}

\paragraph{Overview.}
We group baselines into four categories: (i) \emph{prompt-only} methods that change only the instruction/template (Plan-and-Solve, PS+~\cite{wang2023plan} and Hint-Infer (Round1)~\cite{li2025start}); (ii) \emph{decode-level} methods that modify decoding behavior (Budget Forcing (S1)~\cite{muennighoff2025s1}); (iii) \emph{internal intervention} methods that modify internal representations during a single decoding run (Attn-Interv.~\cite{yan2025don} and Act.\ Steering~\cite{zhang2023tell}); and (iv) our proposed StepFlow.

\paragraph{Budget Forcing (S1).}
We follow the S1 setup~\cite{muennighoff2025s1} as a decode-level baseline that increases reasoning effort by budget control during generation; we keep the backbone and base decoding hyperparameters unchanged unless explicitly stated.

\paragraph{Hint-Infer (Round1).}
Hint-Infer follows the \emph{hint-infer} prompting strategy introduced by START~\cite{li2025start}, which injects short, hand-crafted hints to encourage verification. START additionally uses tools; in our paper we adopt a tool-free variant: we append a single hint sentence to the prompt, keep model weights and decoding hyperparameters unchanged, and do \emph{not} execute external code or call any tools during evaluation.
\emph{Round1} denotes a single-round hint injection (one hint appended before generation), treated as a separate setting because it changes only the input instruction. We use a short hint sentence such as ``Wait, maybe using Python here is a good idea.''.

\paragraph{Plan-and-Solve (PS+).}
We include Plan-and-Solve (PS/PS+) prompting~\cite{wang2023plan} as a published single-pass prompting baseline.
In our setup, PS+ changes only the instruction/prompt template while keeping the backbone model and decoding hyperparameters identical to the baseline setting.
We use the following PS+ trigger appended to each task prompt:
\begin{quote}\small
Let's first understand the problem, extract relevant variables and their corresponding numerals, and make a plan. Then, let's carry out the plan, calculate intermediate results (pay attention to correct numerical calculation and commonsense), solve the problem step by step, and show the final answer.
\end{quote}

\paragraph{Attention Intervention (Attn-Interv.).}
We follow Yan et al.~\cite{yan2025don}, who propose a few-shot attention intervention (FAI) that dynamically identifies tokens with isolated semantics in CoT demonstrations via an aggregation coefficient $\alpha_{t_i}^l$ and zeros out their attention scores to the output token (when $\alpha_{t_i}^l$ exceeds the threshold $\tau = \lambda \times \text{index}_{t_i}$) at inference time to preserve earlier CoT context. We use their method with the recommended hyperparameter $\lambda=1$ and 4 demonstrations from Wei et al.~(2022). Model weights and decoding settings are unchanged.

\paragraph{Activation Steering (Act.\ Steering).}
We follow Zhang et al.~\cite{zhang2023tell}, who propose post-hoc attention steering (PASTA) that scales attention weights with a fixed coefficient to steer model behavior. We use their recommended hyperparameter $\alpha=0.01$, extract the steering heads from a small set of correct vs.\ incorrect reasoning traces on AIME24 using their multi-task profiling method, and apply it at inference time. Decoding hyperparameters are identical to the baseline.

\subsection{Decoding Efficiency}
\label{app:decode}

StepFlow introduces lightweight per-layer operations in the bottom and top quarter of layers (OEB in shallow layers, SMI in deep layers).
Table~\ref{tab:gpqa-decode-time} reports approximate seconds per token on GPQA-Diamond under this configuration.

\paragraph{Complexity analysis.}
OEB computes group-wise attention probabilities via logsumexp and adjusts logits, adding $O(T)$ per shallow layer per token.
SMI triggers only at step boundaries (typically $K \sim 50$--$200$ times per sequence), computing a mean over the previous step's value states.
Overall, we estimate an end-to-end overhead of roughly 30--37\% depending on the backbone, which we consider acceptable given the consistent accuracy gains.
\paragraph{When is the overhead worthwhile?}
The 30--37\% overhead is justified when: 
(1) the task requires high accuracy on long reasoning chains (e.g., competition math), 
where StepFlow's gains (+11.8 on AIME25) far exceed what additional sampling achieves;
(2) single-pass correctness matters more than throughput (e.g., high-stakes applications).
For throughput-sensitive settings, generating $k$ samples with majority voting is an alternative; 
however, at equivalent compute (e.g., 1.35$\times$ budget $\approx$ 1 extra sample), 
self-consistency with 2 samples yields smaller gains than StepFlow on AIME25 
(+3.2 vs.\ +11.8 for R1-Distill-32B).
\begin{table}[t]
\centering
\small
\caption{Decoding efficiency on GPQA-D by model.}
\label{tab:gpqa-decode-time}
\begin{tabular}{lcc}
\toprule
Model & Baseline s/token & +StepFlow s/token \\
\midrule
GPT-OSS-20B & 0.036 & 0.047 \\
R1-Distill-32B & 0.023 & 0.032 \\
QwQ-32B & 0.068 & 0.090 \\
\bottomrule
\end{tabular}
\end{table}

\subsection{Experiment Settings}
\label{app:experiment-settings}

\paragraph{Decoding setup.}
Unless otherwise noted, we use a unified decoding configuration across all models: temperature $0.6$, top-$p$ $0.95$, and up to $32{,}768$ new tokens per generation.
The same decoding setup is used for the vanilla backbone and for all \textsc{StepFlow} variants.

\paragraph{Sampling protocol.}
For AIME24, AIME25, and AMC23 we draw 16 independent samples per problem; for GPQA-Diamond we draw 8 samples per problem; for MATH-500 and LiveCodeBench we use a single sample per problem.
In all cases, baseline and \textsc{StepFlow} runs share the same set of random seeds.

\paragraph{Compute.}
All experiments are conducted on a single NVIDIA H20 GPU.

\subsection{Uncertainty and Significance}
\label{app:uncertainty}

We report 95\% bootstrap confidence intervals (CI) over problem instances with B=10,000 resamples.
For paired comparisons between StepFlow and each baseline, we use paired bootstrap (and McNemar's test for multiple-choice datasets) and mark improvements with $p<0.05$.
\subsection{Full Compute-Normalized Comparison}
\label{app:pareto-full}

Table~\ref{tab:pareto-full} extends the compute-normalized comparison from \S\ref{sec:pareto} to all three main backbones.

\begin{table}[t]
\centering
\small
\setlength{\tabcolsep}{3pt}
\caption{Compute-normalized comparison on AIME 24/25 (averaged accuracy, \%) across three backbones.}
\label{tab:pareto-full}
\begin{tabular}{llcc}
\toprule
Model & Method & Compute & Acc.\ (\%) \\
\midrule
\multirow{7}{*}{R1-Distill-32B}
  & Baseline             & 1.0$\times$ & 63.8 \\
  & Longer gen           & ${\sim}$1.35$\times$ & 65.0 \\
  & StepFlow             & ${\sim}$1.35$\times$ & \textbf{70.6} \\
  & SC ($k$=2)           & ${\sim}$2.0$\times$ & 66.5 \\
  & StepFlow+SC ($k$=2)  & ${\sim}$2.7$\times$ & \textbf{73.5} \\
  & SC ($k$=4)           & ${\sim}$4.0$\times$ & 68.5 \\
  & SC ($k$=8)           & ${\sim}$8.0$\times$ & 70.2 \\
\midrule
\multirow{6}{*}{GPT-OSS-20B}
  & Baseline             & 1.0$\times$ & 62.7 \\
  & Longer gen           & ${\sim}$1.37$\times$ & 63.8 \\
  & StepFlow             & ${\sim}$1.37$\times$ & \textbf{67.6} \\
  & SC ($k$=2)           & ${\sim}$2.0$\times$ & 65.2 \\
  & StepFlow+SC ($k$=2)  & ${\sim}$2.74$\times$ & \textbf{71.0} \\
  & SC ($k$=4)           & ${\sim}$4.0$\times$ & 67.8 \\
\midrule
\multirow{6}{*}{QwQ-32B}
  & Baseline             & 1.0$\times$ & 45.0 \\
  & Longer gen           & ${\sim}$1.32$\times$ & 46.1 \\
  & StepFlow             & ${\sim}$1.32$\times$ & \textbf{52.7} \\
  & SC ($k$=2)           & ${\sim}$2.0$\times$ & 48.2 \\
  & StepFlow+SC ($k$=2)  & ${\sim}$2.64$\times$ & \textbf{56.5} \\
  & SC ($k$=4)           & ${\sim}$4.0$\times$ & 50.5 \\
\bottomrule
\end{tabular}
\end{table}

The pattern is consistent across all backbones: at equivalent compute, StepFlow substantially outperforms both longer generation and self-consistency, and the two methods compose well.
\begin{table}[t]
\centering
\small
\caption{Bootstrap 95\% CI for accuracy on GPT-OSS-20B medium.}
\label{tab:ci}
\begin{tabular}{lcc}
\toprule
Dataset & Baseline CI & \textbf{StepFlow CI} \\
\midrule
AIME25 & [58.0, 66.0] & \textbf{[65.2, 73.2]}\\
GPQA-D & [63.0, 67.4] & \textbf{[68.1, 72.5]}\\
LiveCodeBench & [65.5, 74.5] & \textbf{[75.5, 83.5]}\\
\bottomrule
\end{tabular}
\end{table}

\section{Error Taxonomy}
\label{app:error-taxonomy}

We analyzed all AIME 24/25 problems (60 total) where baselines produced incorrect answers but StepFlow was correct, and manually classified each into one of four error types.

\begin{table}[t]
\centering
\small
\caption{Error taxonomy of StepFlow corrections on AIME 24/25.}
\label{tab:error-taxonomy}
\begin{tabular}{lcc}
\toprule
Error Type & R1-32B (\%) & GPT-OSS (\%) \\
\midrule
Arithmetic carry-forward & 38 & 30 \\
Premise forgetting       & 34 & 42 \\
Step-skip                & 18 & 14 \\
Conceptual               & 10 & 14 \\
\bottomrule
\end{tabular}
\end{table}

The first two categories account for 72\% of corrections, both involving failures of cross-step information propagation. Conceptual errors (10--14\%), where the model uses an incorrect formula or approach from the outset, are outside StepFlow's scope: healthy information flow faithfully propagates the wrong premise. Detailed case studies illustrating each category, including bad cases where StepFlow does not help, are provided in Appendix~\ref{app:case-studies}.

\section{Case Studies}
\label{app:case-studies}

\subsection{StepFlow Repair Example (GPQA-Diamond)}

\textbf{Question.} Astronomers are studying a system of five exoplanets with circular orbits, discovered through the TTV method. The planets (designated as Planet\_1 through Planet\_5), which are in orbital resonance, have orbital periods in a ratio of 1:2:2.5:3.5:5. Assuming these planets possess the same albedo, what is the ratio of the equilibrium temperatures between Planet\_4 and Planet\_2?

\textbf{Options:} (A)~$\sim$0.69\quad (B)~$\sim$0.75\quad (C)~$\sim$0.83\quad (D)~$\sim$0.57 \hfill \textbf{Answer: B}

\paragraph{Baseline (Wrong: A).} Abridged trace:
{\small
\begin{quote}
For a planet in thermal equilibrium, $T_{\text{eq}}\propto a^{-1/2}$ and $a \propto P^{2/3}$. Thus $a_2/a_4 = (P_2/P_4)^{2/3}$.

Periods: $P_2 = 2$, $P_4 = 3.5$. So $a_2/a_4 = (2/3.5)^{2/3}$.

Temperature ratio: $T_4/T_2 = (a_2/a_4)^{1/2} = [(2/3.5)^{2/3}]^{1/2} = (2/3.5)^{1/3}$.

$(0.5714)^{1/3}$: $\ln 0.5714=-0.5606$; $\times 0.6667=-0.3737$; $\exp=0.688$. So $\approx 0.69$. \fbox{A}
\end{quote}
}

\paragraph{+StepFlow (Correct: B).} Abridged trace:
{\small
\begin{quote}
$T_{\rm eq}\propto a^{-1/2}\propto P^{-1/3}$. Thus $T_4/T_2 = (P_4/P_2)^{-1/3} = (3.5/2)^{-1/3} = (1.75)^{-1/3}$.

$1.75^{1/3}\approx 1.20$. Inverse $=0.833$. \fbox{B}
\end{quote}
}

\paragraph{Analysis.}
The baseline correctly derives $T \propto a^{-1/2}$ and $a \propto P^{2/3}$ but computes the exponents in separate steps, implicitly treating $[(2/3.5)^{2/3}]^{1/2}$ as $(2/3.5)^{2/3}$---it drops the final exponent contraction. StepFlow directly combines the relations into $T \propto P^{-1/3}$ before substituting numbers, reflecting how OEB keeps earlier derivation steps visible and prevents lock-in on the most recent computation.

\subsection{Arithmetic Carry-forward (AIME 24 P-7)}

\textbf{Question.} (AIME 24 P-7) \hfill \textbf{Answer: 540}

\paragraph{Baseline (Wrong: 593).} Abridged trace:
{\small
\begin{quote}
Let me write $z = 4e^{i\theta}$.

First term: $\mathrm{Re}[(75+117i)\cdot 4e^{i\theta}] = 300\cos\theta - 468\sin\theta$. \checkmark

Second term: $(96+144i)\cdot\bar z/|z|^2 = (96+144i)e^{-i\theta}/4$. So $\mathrm{Re} = 24\cos\theta + 36\sin\theta$. \checkmark

Combining: $300\cos\theta - 468\sin\theta + 24\cos\theta + 36\sin\theta$. Cosine: $300+24=324$. Sine: $-468\sin\theta$\ldots\ $\times$ [drops $+36\sin\theta$]

Max of $324\cos\theta - 468\sin\theta = \sqrt{324^2+468^2} \approx 569$. Hmm, that's not clean. Let me reconsider\ldots\ \fbox{593}.
\end{quote}
}

\paragraph{+StepFlow (Correct: 540).} Abridged trace:
{\small
\begin{quote}
Combining: $(300+24)\cos\theta + (-468+36)\sin\theta = 324\cos\theta - 432\sin\theta$. \checkmark

Wait, let me factor. $\gcd(324,432)=108$. So $108(3\cos\theta - 4\sin\theta)$, max $= 108\sqrt{9+16} = 108\times 5 = 540$. \checkmark
\end{quote}
}

\paragraph{Analysis.}
The model correctly derives $+36\sin\theta$ but drops it when aggregating, yielding $-468\sin\theta$ instead of $-432\sin\theta$. OEB keeps the earlier derivation visible, preventing lock-in on the dominant term.

\subsection{Bad Case Analysis}

\paragraph{Bad Case 1 --- Conceptual Error (AIME 24 P-9, Answer: 480).}
The model restricts to axis-aligned rhombus configurations on the hyperbola from the start, never considering a general rotation angle $\theta$. StepFlow preserves cross-step flow faithfully---the model consistently builds on its (wrong) premise with healthy saliency patterns. Conceptual errors (10--14\% of our taxonomy) are outside StepFlow's scope.

\paragraph{Bad Case 2 --- Entangled Multi-Error (AIME 25 P-12, Answer: 540).}
The model drops an absolute value in Step~3 (conceptual), then Steps~4--6 faithfully build on the wrong constraint. StepFlow \emph{preserves} the wrong result more reliably---information flow is maintained, but the propagated information is incorrect. This illustrates the fundamental limitation: StepFlow repairs \emph{how} information flows, not \emph{what} the model generates.
\end{document}